# Tell Me Why Is It So? Explaining Knowledge Graph Relationships by Finding Descriptive Support Passages

**Sumit Bhatia$^\alpha$, Purusharth Dwivedi$^\beta$ and Avneet Kaur$^\beta$**
$^\alpha$IBM Research
$^\beta$ IIIT Delhi
sumitbhatia@in.ibm.com, {purusharth14081,avneet14027}@iiitd.ac.in

## Abstract

We address the problem of finding descriptive explanations of facts stored in a knowledge graph. This is important in high-risk domains such as healthcare, intelligence, etc. where users need additional information for decision making and is especially crucial for applications that rely on automatically constructed knowledge bases where machine learned systems extract facts from an input corpus and working of the extractors is opaque to the end-user. We follow an approach inspired from information retrieval and propose a simple and efficient, yet effective solution that takes into account passage level as well as document level properties to produce a ranked list of passages describing a given input relation. We test our approach using Wikidata as the knowledge base and Wikipedia as the source corpus and report results of user studies conducted to study the effectiveness of our proposed model.

## 1 Introduction

Knowledge Graphs are becoming increasingly important in knowledge and data management applications as they afford a semantic structure to the underlying data. They form crucial components of modern web search engines, state-of-the-art question answering systems such as IBM Watson, and are used in a variety of applications in domains as diverse as healthcare (Nagarajan and others, 2015), finance (Ruan et al., 2016), media (Gutirrez-Cuellar and Gmez-Prez, 2014), cybersecurity (Iannacone et al., 2015), etc.

Entities are the fundamental units of knowledge graphs and are often presented to users as a result of a search query, or are used in applications such as exploratory search where users can search about entities of interest and browse their important relationships (Heim et al., 2010). For various critical applications such as exploring interactions between genes and drugs (Fokoue et al., 2016), intelligence applications (Sheth et al., 2005), etc., users may want some additional description or supporting evidence that provides some explanation of the relationship presented to them in order to build confidence in their decision making process. Even for generic information search or browsing activities the entities and relationships presented to the user may be unknown to her and thus, she may not be able to fully appreciate the relevance of information presented to her by the system. As an example, consider the relationship triple <*H. R. McMaster, military_rank, Lieutenant General* > and its following description as extracted by our proposed approach (Section 3).

> *...In February 2014, Defense Secretary Chuck Hagel nominated McMaster for Lieutenant General and in July 2014, McMaster pinned on his third star when he began his duties as Deputy Commanding General of the Training and Doctrine Command and Director of TRADOCs Army Capabilities Integration Center. Army Chief of Staff General Martin Dempsey remarked in 2011 that McMaster was "probably our best Brigadier General. McMaster made Times list of the 100 most influential people in the world in April 2014...*

An end-user who does not know that Mr. McMaster is a US Army officer may find the above fact much more useful when presented with the accompanying supporting explanation rather than presenting the fact alone. It may also help build his trust and confidence in the system. In fact, it has been found that in scenarios where users are dealing with uncertain information, use of natural language descriptions helps in the decision making process (Gkatzia et al., 2016). In web search engines, usefulness of small text snippets to improve end-user experience is well studied (Clarke et al., 2007). Likewise, in context of scientific digital libraries, it has been found that accompanying figures, tables, etc. with small textual descriptions helps users in judging their importance (Sandusky and Tenopir, 2008; Demner-Fushman et al., 2009). Therefore, we posit that providing users with small textual explanations of the relationships may help their understanding, build their confidence in the system and help them in accomplishing their intended tasks. We believe that such a capability is even more crucial for systems that rely on knowledge graphs that are constructed automatically (Niu et al., 2012), especially using deep neural networks (Socher et al., 2013) where interpretability is a big issue.

In this work, we describe *a probabilistic method based on language models to extract supporting passages from an underlying text corpus that provide descriptive explanations of a knowledge graph relationship* (Section 3). Given an input relationship, our model takes into account passage-level and document-level evidence to rank different passages in the order of their relevance to the input relationship. Previous works on explainability of knowledge graph data have mainly focused on explaining how two entities in a graph may be related and the *explanations* are often in a form of a set of common entities or paths connecting the two entities (Fang et al., 2011; Aggarwal et al., 2016; Pirrò, 2015) and thus, suffer from the same issues as discussed above. Efforts on generating textual descriptions of relationships have also focused mainly on template based methods where given a set of facts and an underlying text corpus, different templates are learned that could be used for representing the relationship (Althoff et al., 2015; Voskarides et al., ). For example, for relations of type $<X, dateOfBirth, Y>$, sentence templates such as "X was born on Y" are learned. However, such sentences offer textual *representations* of the input relationship rather than a *supporting explanation* which is the main focus of our work. Further, our proposed approach is simple, effective, and unsupervised, and thus, can be easily adopted by different systems. We implemented and evaluated our approach using Wikipedia as our background text corpus and Wikidata as our knowledge base and results obtained through user studies conducted to study the effectiveness of our proposed techniques are encouraging (Section 4).

## 2 Related Work

We provide a brief overview of related work categorized under two broad categories. First we provide an overview of few representative papers that have looked at generating small textual descriptions of results in different search scenarios such as web search, academic search, etc. Next, we focus on works that have addressed the problem of explaining relatedness between knowledge graph entities through both graphical and textual summaries.

### 2.1 Supporting Search Results With Textual Descriptions

User studies conducted by Tombros and Sanderson (Tombros and Sanderson, 1998) have shown that in document retrieval systems, presenting users with short textual summaries describing the retrieved documents help them judge the importance and utility of the results much better and faster. Likewise, in Web Search Engines, it is a common practice to present results along with a small textual summary or *snippet* extracted from the web page (Turpin et al., 2007) and the positive influence of snippets on end-user experience and behavior is well studied (Clarke et al., 2007). In context of academic search engines such as CiteSeer and Google Scholar, Bhatia and Mitra (Bhatia and Mitra, 2012) studied the problem of generating small descriptions of *document-elements* (figures, tables, and pseudo-codes) present in academic papers to help users quickly decide their importance without having to read the whole paper. Similarly, snippets have been found useful for XML search systems (Huang et al., 2008) and ontology search systems (Penin et al., 2008) where small textual descriptions accompanying search results have helped users select the most suitable result for their information needs out of the presented results.

### 2.2 Explaining Knowledge Graph Relationships

**Graph Based Approaches:** On receiving an entity query, Web search engines such as Google, Bing, etc. often show a list of related entities on the search page or in a separate entity box populated by information derived from the underlying knowledge base. However, it is not always apparent to the users how the suggested entities are connected to the input entity. Fang et al. (Fang et al., 2011) describe their system

*REX* that takes as input two knowledge graph entities and produces a ranked list of relationships between the two entities efficiently. Bhatia et al. (Bhatia et al., 2016) proposed a relationship ranking function that takes into account features such as entity popularity, affinity between the input entities and strength of different relationships between them.

Pirrò (Pirrò, 2015) considered the problem of explaining how two entities in a knowledge graph might be related as a sub-graph finding problem where the sub-graph consists of nodes and edges in the set of paths between the two input entities. Thus, the explanation of the relatedness between two entities is provided by means of shared entities and relationships between them. Likewise, Aggarwal et al. (Aggarwal et al., 2016) considered the task of explaining relationships between two entities as a path-ranking problem and propose a scoring mechanism to identify informative and discriminative paths.

**Text Based Approaches:** In context of web search where the systems present entities as part of search results, Blanco and Zaragoza (Blanco and Zaragoza, 2010) studied the problem of finding support sentences for explaining why an output entity is considered relevant to the original ad-hoc text query by the user.

Saldanha et al. (Saldanha et al., 2016) addressed the problem of generating descriptions of lesser known companies and describe a template based approach to create such descriptions by generating sentences from RDF triples found in DBPedia and Freebase about the company. These sentences are generated by utilizing the RDF triples and corresponding Wikipedia sentences for known companies and learning templates such as "$< company >$ was founded by $< founder >$". Voskarides et al. (Voskarides et al., 2015) describe a learning to rank based sentence extraction and ranking method to find human readable descriptions of a relationship between two knowledge graph entities. Their follow-up work (Voskarides et al., ) tackles the problem using a template based approach. For a given relationship type, they identify representative sentences describing some of the relationship instances and then generating textual description of other instances of the same relationship type by selecting a suitable template and filling it with appropriate entities.

Such template based approaches requires manual construction of templates for each relationship type that may be difficult for many practical applications. For example, Wikidata contains more than 1600 unique relationships types, DBPedia contains more than 2800 relationship types. The problem is exacerbated in domain specific knowledge graphs where domain knowledge is required for generating appropriate templates. Further, machine learning of such templates or other learning based methods require significant amount of training data and it may not always be feasible due to lack of such data and thus, may only be useful for a few specific relationship types.

## 3 Proposed Approach

Let us consider a relationship $\mathcal{R} =< s, r, t >$ in a knowledge Graph $\mathcal{K}$ where $s$ and $t$ correspond to the source and tail nodes (entities), respectively, and $r$ is the relationship edge label. Let $P$ be the set of passages extracted from an underlying text corpus[1]. We wish to rank the passage $p \in P$ based on the probability that it contains a descriptive explanation of $\mathcal{R}$. Probabilistically, we are interested in computing the probability that the relationship $\mathcal{R}$ is mentioned in passage $p$, i.e., $P(p|\mathcal{R})$.

By application of Bayes' Theorem, we have:

$$P(p|\mathcal{R}) = \frac{P(p) \times P(\mathcal{R}|p)}{P(\mathcal{R})} \propto P(p) \times P(\mathcal{R}|p) \quad (1)$$

Here, $P(\mathcal{R})$ in the denominator has been ignored as it will be same for all the passages $p \in P$. The component $P(p)$ can be interpreted as the prior probability of the passage $p$ being of interest. Note that this prior is independent of the relationship (query) and can be used to model certain domain specific characteristics based on the application requirements. For example, in a medical domain application, passages coming from a peer-reviewed article can be assigned a higher prior than passages coming from a non-authoritative article. In this work, we are focused on the general performance of the framework and hence, we assume a uniform prior as is common in information retrieval (Manning et al., 2008, Chapter 12) and thus, $P(p)$ can also be ignored for ranking purposes. With these assumptions and assuming conditional independence of $s, r,$ and $t$, equation 1 reduces as follows.

---

[1]Given a text corpus, there are multiple ways of extracting passages and the approach for ranking these passages is independent of the way passages were created. We detail our choice of passage extraction method in the section on experiments (Section 4).

$$P(p|\mathcal{R}) \propto \underbrace{P(s|p) \times P(t|p)}_{\text{entity probability}} \times \underbrace{P(r|p)}_{\text{relationship probability}} \qquad (2)$$

Here, $P(s|p)$ and $P(t|p)$ represent the probability of observing mentions of source and tail entities, $s$ and $t$, respectively in the passage $p$. Likewise, $P(r|p)$ represents the probability that relation label $r$ is being described in passage $p$. In order to compute these probabilities, we adapt the query likelihood model based on multinomial unigram language model (Manning et al., 2008) that computes probability of generating a query given a text document. We can treat each passage in $P$ as our source document and compute the probabilities of generating the entities $s, t$ and relation $r$ as specified in equation 2. Note that the names of entities $s$ and $t$ and relationship label $r$ consist of multiple individual words and assuming conditional independence of terms, we can simplify equation 2 as follows.

$$P(p|\mathcal{R}) \propto \prod_{w \in S \cup T \cup R} P(w|p), \qquad (3)$$

Here, $S$, and $T$ are the sets of terms in names of source entity $s$ and target entity $t$, respectively, and $R$ is the set of terms representing the relationship $r$. Note that relationship labels in knowledge graphs are often created like variable names (*bornOn, citizen_of*, etc.) that are generally not used in standard written vocabulary. Further, a given relationship may be described by different synonymous terms (occupation, profession, etc.). Therefore, to account for these variations, $R$ can be constructed by using a set of synonyms representing a given relationship type. In this work, we have chosen relationship label aliases provided by Wikidata to obtain a set of terms that could be used for representing a given relationship type. Depending upon the application at hand, different domain specific synonyms can also be used for this purpose.

Another important consideration is that a typical passage is only a few sentences long. As a result, a given passage alone may not have sufficient information to reliably approximate the probability of observing a term from the passage due to data sparsity issues. The probabilities are over estimated for the terms that are present in the passage and are under estimated for the terms that are not present in the passage. This is especially exacerbated in case of entity names (nouns) that are often mentioned as corresponding pronouns (his, her, she, etc.). As a result, a highly useful passage may get a very low score if the entity of interest is mentioned by its pronoun in the passage. In order to account for such imbalances, the probability estimations are smoothed by adding document and collection level statistics. Consequently, the unigram language model of passage $p$ is then modeled as a mixture of passage, document, and collection (corpus) language models, respectively, as follows:

$$P(w|p) = P(w|\Theta_{MM}) \qquad (4)$$
$$= \lambda_1 \underbrace{P(w|\Theta_p)}_{\text{passage-level evidence}} + \lambda_2 \underbrace{P(w|\Theta_d)}_{\text{document-level evidence}} + \lambda_3 \underbrace{P(w|\Theta_c)}_{\text{collection-level evidence}} \qquad (5)$$

where, $\lambda_1 + \lambda_2 + \lambda_3 = 1$. We set $\lambda_1 = \lambda_2 = 0.4, \lambda_3 = 0.2$ for our experiments.

Modeling the entity probabilities and smoothing as just described serves multiple objectives. First, it helps overcome the sparsity problem due to the short length of the passage. Second, the document level evidence gives a higher score to passages that come from documents that talk more about the entities involved in input relationship. Thus, passages coming from documents that are majorly about the involved entities are given a higher weight by the ranking function described in equation 5. Also note that such a formulation also addresses the problem of co-reference resolution (Hajishirzi et al., 2013) to some extent and can be interpreted as a probabilistic variant of the heuristic used by Wu and Weld (Wu and Weld, 2010) that replaces most frequent pronouns in Wikipedia article with article title. Lastly, the collection level evidence is also important as it plays the role of a reference or *background* language model and provides term weighing similar to inverse document frequency (IDF) (Zhai and Lafferty, 2001).

The individual probabilities in equation 5 can be computed by using the statistics from passage, document, and collection as follows:

$$\text{Passage Evidence: } P(w|\theta_p) = \frac{count(w,p) + 1}{|p| + |V|} \qquad (6)$$

$$\text{Document Evidence: } P(w|\theta_d) = \frac{count(w,d) + 1}{|d| + |V|} \qquad (7)$$

$$\text{Collection Evidence: } P(w|\theta_c) = \frac{count(w,c)}{|V|} \qquad (8)$$

Here, $V$ is the vocabulary of the corpus and $|\cdot|$ indicates the size of the set. Note that we have added the constant one in equations 6 and 7 to prevent zero probabilities for terms that may not be present in the respective passage or document. Further, the denominators are chosen so that the sum of probabilities over the entire vocabulary is one. Also note that the additive factor is not required in the collection model as all the terms in the vocabulary are present in the collection by definition.

## 4 Experimental Evaluation

### 4.1 Data Description

In this section we discuss the dataset used in our experiments and how the queries and relevance judgments were obtained.

#### 4.1.1 Relationship Queries:

We need relationship triples of form $<s, r, t>$ that will constitute our query relationships for which the supporting passages need to be retrieved from the underlying corpus. In order to create such a query set, we selected title of top 25 most viewed pages[2] for the months of January-April, 2017. From these page titles, we retained only those titles that correspond to named entities by manually filtering out titles like *List of Black Mirror episodes, Deaths in 2017*, etc. That gave us a total of 80 unique entities. Next, we used Wikidata[3] as our knowledge base and retrieved all relationships of the entities selected previously using the SPARQL end-points provided by Wikidata. From all these retrieved relationships, we filtered out relationships that were not in English language, were of type *instance of* and *subclass of*, and where the target entity was not a named entity. This resulted in a final set of 1250 unique relationship triples from which we selected 150 triples at random as our final relationship query set that was used in subsequent experiments.

#### 4.1.2 Source Corpus and Passages:

We chose Wikipedia[4] as our underlying corpus. There are multiple ways to extract a set of passages given a text corpus such as utilizing the document structure and paragraph or section markers present in the documents itself. However, the passages thus extracted are usually very long, often running into tens of sentences. Further, while such paragraph or section markers are available for well structured corpora such as Wikipedia, they may not always be available for different source documents. More importantly, such long passages may be detrimental to the end-user experience as they consume valuable screen real estate and reading them requires significant additional efforts from users. Another option is to use text segmentation methods such as TextTiling (Hearst, 1997) that segment the input text into topically coherent passages. However, such approaches require significant pre-processing efforts, especially for large corpora (few millions of documents) often encountered in real world applications. In practice, simple (and fast) segmentation of input text into fixed length, overlapping passages using a sliding window approach is found to be equally effective (Tiedemann and Mur, 2008; Tiedemann, 2007; Khalid and Verberne, 2008), if not better, and is the approach we also take. Use of overlapping passages is also encouraged as it reduces the chances of relevant information getting split between two consecutive passages (Callan, 1994). Therefore, we split the input text of each document into overlapping passages of three consecutive sentences using a sliding window of size three as suggested by Spangler et al. (Spangler et al., 2003). This resulted in about $80.5$ million extracted passages that constitute our source set of passages (set $P$ in Section 3).

---

[2]https://en.wikipedia.org/wiki/Wikipedia:Top_25_Report
[3]https://www.wikidata.org/wiki/Wikidata:Main_Page
[4]Specifically, we used the dump of $20^{th}$ April, 2017.

In order to compute the different passage, document, and collection based statistics, we used the Indri toolkit provided by the Lemur project[5]. The toolkit offers capabilities to query and index a collection of documents, and APIs to query term statistics required for language model based computations described in our ranking function (equation 5). Specifically, we created two indexes using Indri – a *passage index* of all the extracted passages to compute passage level statistics and an *article index* of all the Wikipedia articles (about 5.34 million articles) to compute document and collection level statistics. A standard stopword list provided by the Onix text retrieval toolkit[6] was used to filter out common stop words and stemming was performed using Porter's Stemmer.

### 4.1.3 Baseline:

We used the inference network based generative passage retrieval algorithm implemented in Indri (Metzler and Croft, 2004) as our baseline method. This is a state-of-the-art passage retrieval method and is often chosen as a baseline for various research tasks related to passage retrieval (Wang and Si, 2008; Yang et al., 2006). Given a query, this method finds documents that are relevant for the query and then extracts specific continuous portions of text from the documents that are highly relevant for the query. Given an input relationship tuple $<s, r, t>$, the input query to Indri consists of all the terms in source and target entity names and relationship description. For comparison purposes, we set the length of passages to be returned by Indri as 600 words as this is the average length of passages extracted as described above.

## 4.2 Effectiveness Evaluation

In order to study the effectiveness of our proposed approach to find high quality descriptive passages, we selected a random set of 50 relationship triples from the set of 150 triples described above. For each of these 50 triples, we computed top five passages from the corpus ranked by our ranking function (equation 5). We also obtained five passages for each of the relationship triple by the baseline method as provided by Indri toolkit.

Next, we took help of two human evaluators to evaluate the quality and correctness of the passages retrieved by the baseline and our proposed method. The evaluators were advanced graduate students in Computer Science, not associated with project, and had good command of the English language.

For 50 queries used in this study, say $Q1 - Q50$, first 30 queries ($Q1 - Q30$) were evaluated by one evaluator and last 30 ($Q21 - Q50$) were evaluated by the second evaluator. This way, each evaluator provided judgments for 30 queries and for 10 queries, judgments from both the evaluators were available that were used to study the inter-annotator agreement between them. For each query, all the ten extracted passages for that query (5 each from the baseline and the proposed method) were presented to the evaluators in a randomized order and the evaluators were not informed which passage was retrieved by which method. They were asked to rate each passage on three point scale – 0 if the passage is incorrect, irrelevant or not at all useful, 1 if the passage contains the relationship but is only partially relevant and does not provide a good explanation, and 2 if the passage is correct and highly relevant and provides a good explanation.

---

[5] https://www.lemurproject.org/
[6] http://www.lextek.com/manuals/onix/stopwords1.html

|  | | Evaluator 1 | | |
|---|---|---|---|---|
|  | | **0** (irrelevant) | **1** (partially relevant) | **2** (highly relevant) |
| **Evaluator 2** | **0** (irrelevant) | 25 | 2 | 4 |
|  | **1** (partially relevant) | 9 | 6 | 1 |
|  | **2** (highly relevant) | 7 | 5 | 41 |

Table 1: Distribution of the labels assigned by the two evaluators. Numbers along the diagonals represent cases where the evaluators were in perfect agreement.

|  | P@1 | Precision | MRR |
|---|---|---|---|
| **Baseline** | 0.32 | 0.156 | 0.387 |
| **Proposed Approach** | 0.82 | 0.711 | 0.777 |

Table 2: Performance of the baseline and proposed approach as measured by precision, P@1, and MRR.

#### 4.2.1 Inter Annotator Agreement:

We used Cohen's Kappa coefficient to measure the agreement between the two evaluators. For the 10 queries evaluated by both the evaluators, observed agreement between the evaluators is $0.72$ and the probability of them agreeing at random is $0.39$. This yields the value of Kappa coefficient as $0.54$, indicating moderate to good agreement. Table 1 provides further details about the judgments provided by the two evaluators. 72 out of 100 times, the two evaluators were in perfect agreement (values along the diagonal), 6 times they did not agree on the degree of relevance of a given passage, and 22 times a passage judged as partially or highly relevant by one evaluator was rated as irrelevant by the other evaluator.

Next, in order to study the performance of the two methods, we created a final dataset as follows. For queries $Q1 - Q20$, evaluations provided by annotator 1 were used and for $Q31 - Q50$, evaluations provided by annotator 2 were used. For queries $Q21 - Q30$, for which ratings from both evaluators were present, passages for which the two evaluators were not in perfect agreement, we assigned that passage lower of the two ratings. For example, if a passage was rated as 0 by one evaluator and 1 by the other, we chose 0 as the final rating for that passage. This final merged dataset is then used to compare the performance of the baseline and proposed approach.

Table 2 compares the two approaches by using precision, precision at rank 1 ($P@1$), and mean reciprocal rank (MRR). While precision measures how many of the passages extracted by each method are relevant, $P@1$ and MRR measure the ability of the methods to identify a relevant passage as the top-ranked passage. This is important because in real world applications, due to limited screen real estate and to minimize users' efforts, we want to present the best results at the top position. Note that for the numbers reported in Table 2, only the passages marked as highly relevant (2) by the evaluators were considered as relevant passages. As can be observed, the proposed approach achieves a precision of $0.711$ compared to $0.156$ for the baseline approach. Similar significant out-performance is observed for both $P@1$ and MRR for the proposed approach indicating that compared to the baseline approach, the proposed method is able to find more relevant passages, especially at top ranks.

Next, for a fine-grained analysis, Table 3 provides the distribution of passages marked as irrelevant, partially relevant, and highly relevant for the two approaches. We note that for the proposed approach, only about $15\%$ of the passages were found to be irrelevant by the evaluators compared to about $75\%$ for the baseline approach.

|                 | No. of passages marked as |                   |                 |
|-----------------|:-------------------------:|:-----------------:|:---------------:|
|                 | irrelevant                | partially relevant | highly relevant |
| Baseline        | 188                       | 23                | 39              |
| Proposed Method | 39                        | 32                | 175             |

Table 3: Distribution of judgment labels for the baseline and proposed approach. Note that the total number of passages for proposed approach is 246 instead of 250 because some passages appeared for more than one query.

## 4.3 Preference Evaluation

In this section, we describe the experiment conducted to study the preferences of end-users when passages extracted by the two approaches are presented to them side by side. For this experiment, we used the set of 150 relationship tuples (Section 4.1.1) and recruited 5 undergraduate computer science students that were not associated with this project. For each query, the top scored passage extracted by the baseline and proposed approach was presented to the evaluator side by side and they were asked to select the passage that offered a better description of the relationship tuple. The order in which the passages were presented to the evaluators was randomized and they were not informed of the method that produced a specific passage. Each evaluator provided preference judgments for 30 relationships. The results are summarized in Table 4. As can be seen from the results, for an overwhelming majority of the time, all the evaluators preferred the passages extracted by the proposed approach.

|             | **Baseline** | **Proposal** |
|-------------|:--------:|:--------:|
| **Evaluator 1** | 5        | 25       |
| **Evaluator 2** | 5        | 25       |
| **Evaluator 3** | 5        | 25       |
| **Evaluator 4** | 4        | 26       |
| **Evaluator 5** | 5        | 25       |
| Total       | 24       | 126      |

Table 4: Preferences provided by five evaluators when presented with top passages from the baseline and proposed approach side by side.

## 4.4 Some Illustrative Examples and Error Analysis

In this section, we provide some representative examples to illustrate the strengths and weaknesses of our proposed approach. Consider the relationship <*John Cena, nickname, The Protoype*>, for which the passages as produced by the baseline and our proposed approach are as follows.

> **Baseline:** *A prototype is something that is representative of a category of things, or an early engineering version of something to be tested. Prototype may also refer to: Automobiles. Citron Prototype C, a range of vehicles created by Citron from 1955 to 1956 Citron Prototype Y, a project of replacement of the Citron Ami studied by Citron in the early seventies Daytona Prototype, a sports ca*

> **Proposed approach:** *In 2001, Cena signed a developmental contract with the WWF and was assigned to its developmental territory Ohio Valley Wrestling (OVW). During his time there, Cena wrestled under the ring name The Prototype and held the OVW Heavyweight Championship for three months and the OVW Southern Tag Team Championship (with Rico Constantino) for two months. Throughout 2001, Cena would receive four tryouts for the WWF main roster, as he wrestled multiple enhancement talent wrestlers on both WWF house shows and in dark matches before WWF television events.*

Note that the first passage contains multiple occurrences of the word *prototype* which is also a less frequent word in the corpus, and thus was highly ranked by the baseline approach. On the other hand, the passage produced by the proposed approach is able to correctly identify a good passage even though it only had one occurrence of *prototype*. One reason for this passage getting a very high score is the document level component of the ranking function (equation 5). This passage comes from the Wikipedia article about *John Cena* and thus, its score was boosted by the document-evidence component.

**Error Analysis:** By analyzing the passages extracted by the proposed approach and feedback from the evaluators, we observed two major characteristics of the passages that were not rated as relevant by the evaluators. In the first category, the extracted passage does talk about the entities involved, but it does not provide any description of the relationship specified in the query. Consider the following passage for the relationship <*Alan Comes, employer, Fox News*>.

> ...Goldlines television advertising includes cable networks such as CNN, CNBC, Fox News, History International and Fox Business. Goldline has also been the sponsor of the shows of a number of conservative radio and television hosts, including The American Advisor, and The Glenn Beck Program, The Laura Ingraham Show, The Fred Thompson Show, The Huckabee Report, The Lars Larson Show, The Monica Crowley Show, The Mark Levin Show, and The Alan Colmes Show. In 2009, Goldline incorrectly labeled Glenn Beck as a paid spokesman on its website which raised concerns with his employer, Fox News, which prohibit such a relationship; they later corrected it to radio sponsor...

This passage got a high score by the proposed scoring function because it talks about Fox News and Alan Comes and the originating document also has other mentions of Fox News. However, it does not provide any description about employment of Alan Comes at Fox News. Instead, it provides a lot of unnecessary information to the user.

The other type of passages that were not judged relevant by the evaluators were where there was an indirect reference to the relationship query. Consider the following passage for the query <*Warren Beatty,occupation,Film Producer*>.

> In 1994, Astin directed and co-produced (with his wife, Christine Astin) the short film Kangaroo Court, which received an Academy Award nomination for Best Live Action Short Film. Astin continued to appear in films throughout the 1990s, including the Showtime science fiction film Harrison Bergeron (1995), the Gulf War film Courage Under Fire (1996), and the Warren Beatty political satire Bulworth (1998). After The Goonies, Astin appeared in several more films, including the Disney made-for-TV movie, The B. R. A

Here again, the passage contains a lot of unnecessary information and only contains a fleeting reference to Warren Beatty and the movie Bulworth. There is no explicit mention here that Warren Beatty is a film producer and thus, the evaluators did not find this passage to be very informative.

## 5 Conclusions and Future Work

We studied the problem of providing descriptive explanations for relationships in a knowledge graph and described a probabilistic method for ranking passages derived from an input corpus in order of their relevance to the input relationship. The proposed method is simple, effective, and outperformed state-of-the-art baseline method in user studies conducted for evaluating the effectiveness of proposed approach. One direction for future work is to combine the proposed approach with existing models for graph based explanations for entity relatedness and path ranking and offer textual descriptions for how two entities in the knowledge graph may be related. Such techniques will be useful for discovery and exploratory search based applications and may improve end-user experience by offering human readable explanations of systems' graphical output.